\documentclass[10pt,twocolumn,letterpaper]{article}

\usepackage[pagenumbers]{cvpr} %

\usepackage{tabularx}
\usepackage{multirow}

\definecolor{cvprblue}{rgb}{0.21,0.49,0.74}
\usepackage[pagebackref,breaklinks,colorlinks,allcolors=cvprblue]{hyperref}

\def\paperID{****} %
\def\confName{CVPR}
\def\confYear{2025}

\usepackage{xspace}
\usepackage[export]{adjustbox}
\usepackage{pgf}
\usepackage{xparse}
\usepackage{enumitem}
\usepackage{tikzsymbols}

\usepackage{bold-extra}

\makeatletter

\tikzoption{canvas is xy plane at z}[]{%
  \def\tikz@plane@origin{\pgfpointxyz{0}{0}{#1}}%
  \def\tikz@plane@x{\pgfpointxyz{1}{0}{#1}}%
  \def\tikz@plane@y{\pgfpointxyz{0}{1}{#1}}%
  \tikz@canvas@is@plane
}

\newcommand\notsotiny{\@setfontsize\notsotiny\@vipt\@viipt}

\makeatother

\newcommand{\ourDataset}{ViCaS}
\newcommand{\ourBaseline}{Video-LLaVA-Seg}

\newcommand{\totalVideosNew}{20,416}

\newcommand{\PAR}[1]{\vskip4pt \noindent {\bf #1.~}}

\definecolor{better_blue}{HTML}{4285f4}

\newtoggle{comments}
\toggletrue{comments}

\iftoggle{comments}{%
\newcommand{\ali}[1]{\textcolor{blue}{\textbf{Ali: }{#1}}}
\newcommand{\jay}[1]{\textcolor[rgb]{0,0.5,0}{\textbf{Jay: }{#1}}}
\newcommand{\xueqing}[1]{\textcolor{brown}{\textbf{Xueqing: }{#1}}}
\newcommand{\ext}[1]{\textcolor{orange}{\textbf{Ext. Rev: }{#1}}}

}
{%
\newcommand{\ali}[1]{}
\newcommand{\jay}[1]{}
\newcommand{\xueqing}[1]{}
\newcommand{\ext}[1]{}

}

\newcolumntype{Y}{>{\centering\arraybackslash}X}

\usepackage{pifont}%

\newcommand{\cmarkg}{\textcolor{c_green}{\ding{51}}}%
\newcommand{\xmarkr}{\textcolor{c_red}{\ding{55}}}%

\definecolor{c_red}{HTML}{ea4335}
\definecolor{c_blue}{HTML}{4285f4}
\definecolor{c_orange}{HTML}{ff6d01}
\definecolor{c_green}{HTML}{34a853}
\definecolor{c_purple}{HTML}{9900ff}
\definecolor{c_other}{HTML}{000000}

\title{\ourDataset{}: A Dataset for Combining Holistic and Pixel-level Video Understanding using Captions with Grounded Segmentation}

\author{
Ali Athar%
\quad
Xueqing Deng%
\quad
Liang-Chieh Chen\\[5pt]
ByteDance Inc.\\[5pt]
\url{https://ali2500.github.io/vicas-project/}\\[5pt]
{\tt\small \{ali.athar,xueqingdeng\}@bytedance.com}\\
}

\begin{document}
\maketitle

\begin{abstract}
Recent advances in multimodal large language models (MLLMs) have expanded research in video understanding, primarily focusing on high-level tasks such as video captioning and question-answering. 
Meanwhile, a smaller body of work addresses dense, pixel-precise segmentation tasks, which typically involve category-guided or referral-based object segmentation. 
Although both directions are essential for developing models with human-level video comprehension, they have largely evolved separately, with distinct benchmarks and architectures. 
This paper aims to unify these efforts by introducing \ourDataset{}, a new dataset containing thousands of challenging videos, each annotated with detailed, human-written captions and temporally consistent, pixel-accurate masks for multiple objects with phrase grounding. 
Our benchmark evaluates models on both holistic/high-level understanding and language-guided, pixel-precise segmentation.
We also present carefully validated evaluation measures and propose an effective model architecture that can tackle our benchmark.

\end{abstract}
    
\section{Introduction}
\label{sec:intro}

\begin{figure}
    \centering
    \includegraphics[width=\linewidth]{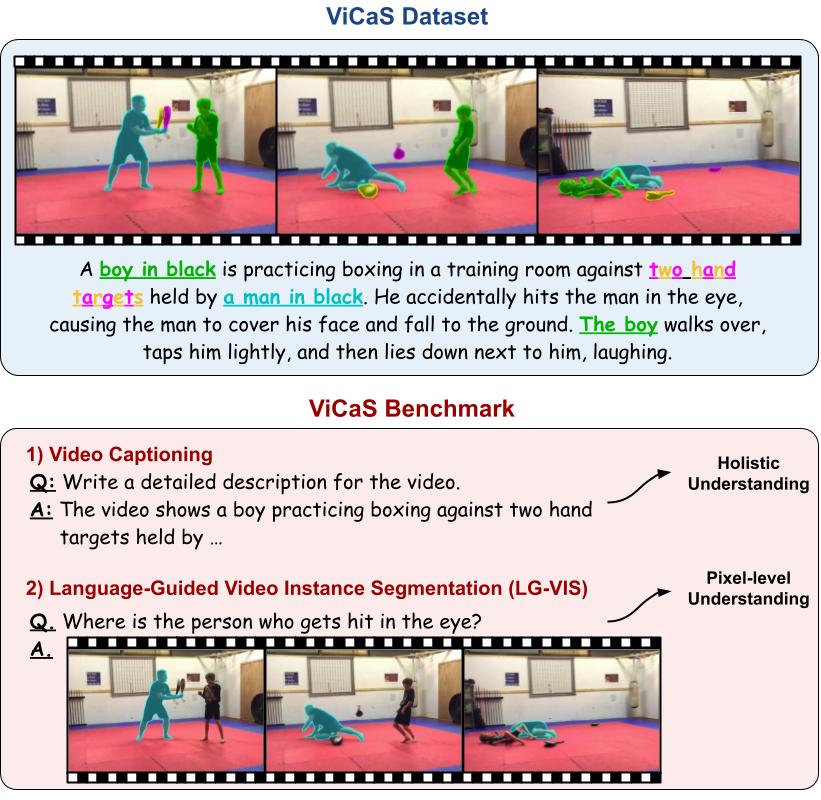}
    \caption{\textbf{\ourDataset{} Dataset/Benchmark.} Our dataset contains detailed video captions with phrase-level grounding for accurate object segmentation masks. The benchmark comprises two tasks to evaluate holistic and pixel-level video understanding, respectively.}
    \label{fig:intro_teaser}
\end{figure}

The emergence of open-source large lauguage models (LLMs)~\cite{dubey2024llama3,bai2023qwen,team2024gemma}, coupled with progress in aligning vision and language feature spaces~\cite{radford2021clip,li2022blip,li2019visualbert}, has enabled significant research into vision-language models that can jointly reason over visual and language inputs.
Earlier advances in image-language models~\cite{liu2023llava,li2023blip2,awadalla2023openflamingo} have spurred development of multimodal LLMs capable of reasoning over other modalities such as videos~\cite{lin2023videollava} and 3D point-clouds~\cite{zhang2022pointclip}.%

In the field of video-language models, research has mainly focused on holistic video understanding tasks such as video captioning and question-answering~\cite{maaz2023videochatgpt,lin2023videollava,zhang2023videollama,xu2024pllava,liu2024llavanext,luo2023valley}, with a particular emphasis on generating temporally dense text output for long video sequences~\cite{zhou2024streaming_vid_cap,song2024moviechat,li2025llamavid,chen2024videollm,yang2023vid2seq}. 
By contrast, comparatively less attention has been given to predicting spatially dense outputs — such as bounding boxes or segmentation masks — that can localize key objects and actors in videos based on text prompts, even though this has important practical applications, \eg, autonomous robots and video editing.
Although recent language-driven benchmarks~\cite{khoreva2019davis_rvos,seo2020ytvos_rvos,ding2023mevis,yan2024visa} in the video segmentation community have made progress in this area, these datasets are largely object-centric and lack evaluation of high-level, holistic video understanding.

In this paper, we seek to bridge the gap between benchmarks that evaluate holistic video understanding and those focused on pixel-level localization.
We introduce \ourDataset{}, a new video dataset and benchmark containing thousands of videos with detailed, human-annotated captions. 
In these captions, words and phrases corresponding to key objects are grounded in human-drawn, pixel-precise segmentation masks that span the entire video duration.
To the best of our knowledge, this is the first video dataset to offer human-labeled annotations of this kind.

Our benchmark comprises two tasks: (1) Video Captioning, which requires describing the video events and objects in detail, and (2) our newly proposed Language-Guided Video Instance Segmentation (LG-VIS) task, which requires predicting temporally-consistent segmentation masks for multiple objects based on a text prompt. 
An example of our dataset annotations and tasks is given in Fig.~\ref{fig:intro_teaser}. 
To effectively evaluate Video Captioning, we conduct a comprehensive user study to validate the evaluation measures used for open-ended text similarity by comparing several existing approaches. 
Finally, we introduce \ourBaseline{}, an end-to-end architecture designed to effectively tackle our benchmark by integrating insights from recent vision-language models~\cite{lin2023videollava,liu2024llavanext} and prompt-based video segmentation approaches~\cite{ravi2024sam2,lai2024lisa}.
In summary, our contributions are as follows:
\begin{itemize}
    \item We introduce a large, human-annotated video dataset with detailed text captions with phrase-level grounding for objects, accompanied by pixel-precise segmentation masks. This dataset enables the evaluation of both holistic and pixel-level video understanding.

    \item We propose accurate, reproducible evaluation measures for open-ended text similarity which are verified by a comprehensive user study.

    \item We present \ourBaseline{}, an effective, end-to-end trained architecture that can tackle our benchmark.
\end{itemize}

\section{Related Work}
\label{sec:related_work}

Although video understanding is a well-researched topic in computer vision, researchers have traditionally approached holistic and pixel-level video understanding as separate streams, each with its own datasets and benchmarks. Below, we review related work in these areas.

\subsection{Holistic Video Understanding}

\vskip-4pt
\PAR{Video Classification}
This is one of the earliest tasks in video understanding, popularized by activity recognition datasets~\cite{kuehne2011hmdb,yu2019activitynet,soomro2012ucf101}, and later expanded by larger datasets~\cite{goyal2017something,sigurdsson2016charades,gu2018ava}, \eg, Kinetics~\cite{kay2017kinetics}. Early deep learning approaches employed 3D CNNs~\cite{karpathy20143dcnn_early,tran20153dcnn1,tran20183dcnn2,tran20193dcnn3,hara20183dcnn4} to capture spatio-temporal information. With the popularization of transformers~\cite{vaswani2017attention}, attention-based architectures~\cite{piergiovanni2023tubevit,arnab2021vivit,liu2022videoswin} emerged, offering improved performance. Although these datasets and architectures have greatly advanced video understanding, they are limited from a language perspective, typically assigning a single, predefined label to each video.

\PAR{Video Captioning and Question-Answering}
Alongside classification, language-oriented tasks such as video captioning and question-answering (Q/A) have gained research attention. Early datasets like MSVD~\cite{chen2011msvd}, MSR-VTT~\cite{xu2016msrvtt}, and TGIF-QA~\cite{li2016tgif} laid the groundwork for video captioning and were later adapted as Q/A benchmarks~\cite{xu2017_msvd_msrvtt}. Initial approaches~\cite{donahue2015RecurrCNN_visRecog,pmlr-v37-showandtell_img_vid_cap} combined CNNs for visual reasoning with RNNs for text generation. As transformers gained popularity, architectures like VideoBERT~\cite{sun2019videobert} and UniVL~\cite{luo2020univl} were among the first to unify vision and language by learning shared representations for both modalities.

\PAR{Multimodal Large Language Models (MLLMs)}
The recent popularization of LLMs~\cite{dubey2024llama3,vicuna2023,bai2023qwen,team2024gemma} has enabled research on multimodal models which extend LLMs to process visual inputs such as images~\cite{liu2023llava,beyer2024paligemma,chen2024vitamin,deng2024coconut,yu2024towards} and videos~\cite{maaz2023videochatgpt,lin2023videollava,song2024moviechat,li2025llamavid,chen2024videollm,luo2023valley}. This research has been supported by large-scale video captioning datasets~\cite{xue2022hd-villa-100m,bain2021webvid10m,chen2024panda70m} and multi-task video understanding benchmarks~\cite{li2024mvbench,cores2024tvbench}.

Overall, the datasets and models discussed above focus primarily on high-level, holistic video understanding and do not address finegrained, pixel-level localization.

\subsection{Pixel-level Video Understanding}

\vskip-4pt
\PAR{Object Tracking and Segmentation}
Object localization and tracking is a deeply studied problem, even prior to the deep learning era~\cite{paragios2000tracking_predl,andriluka2008mot_tracking_people}. This research has been propelled by several datasets, with early efforts focused on bounding-box-level object tracking~\cite{dendorfer2021motchallenge,geiger2012kitti,yu2020bdd100k,fan2019lasot,VOT_TPAMI,dave2020tao,Huang2021got10k}. As model architectures advanced, benchmarks for pixel-precise video segmentation emerged, covering tasks such as object segmentation based on predefined categories~\cite{voigtlaender2019mots,yang2019youtubevis,kim2020video,qiao2021vip,weber2021kittistep,qi2022ovis,miao2022vipseg,mei2022waymo,athar2023burst} or segmenting specific objects given their first-frame ground-truth masks~\cite{pont2017davis,xu2018youtubevos,ding2023mose} or points~\cite{athar2023burst,zulfikar2024pointvos,homayounfar2021videoclick}. Popular approaches have evolved from tracking-by-detection methods~\cite{luiten2018premvos,voigtlaender2019mots,tang2017mot_lmp,braso2020mot_neural_solver} to end-to-end trainable transformer-based~\cite{wang2020visTr,wu2022idol,kim2022tubeformer,athar2023tarvis,shin2024videokmax,zhang2023dvis++,miran2022vita,he2023maxtron}, and object-level attention-based architectures~\cite{cheng2021stcn,cheng2022xmem,oh2019stmvos,athar2022hodor,cheng2024cutie}. Despite substantial progress, these works focus solely on tracking and segmentation, without addressing holistic video understanding.

\begin{figure*}[t]
    \centering
    \includegraphics[width=\textwidth]{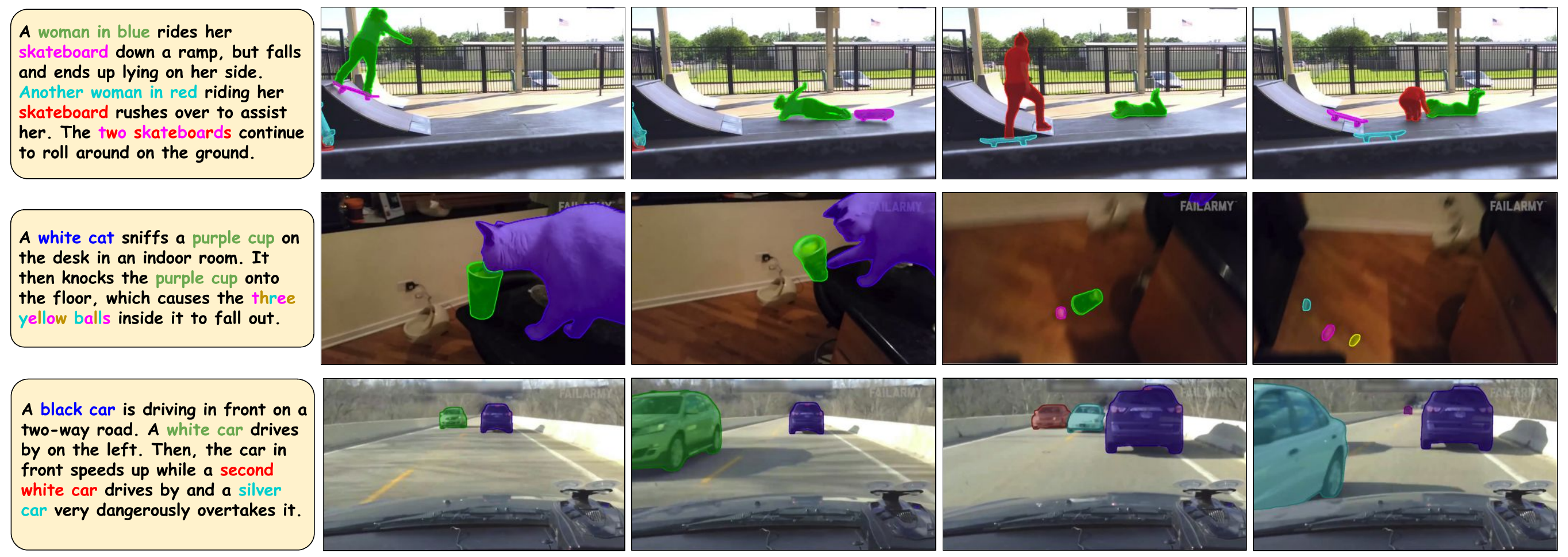}
    \caption{\textbf{\ourDataset{} Examples.} Our dataset showcases diverse scenes with a variety of objects and video events, along with detailed captions. 
    Phrases referring to multiple objects are written with multiple colors, \eg, ``three yellow balls'' in row 2 references three different objects.
    }
    \label{fig:dataset_examples}
\end{figure*}

\PAR{Language-Guided Segmentation}
Recent progress in vision-language models has motivated multiple language-guided video segmentation datasets~\cite{voigtlaender2023vln}. These include ``open-vocabulary'' datasets~\cite{wang2023lvvis,wang2021uvo} that cover a large set of object classes, and referral-based video segmentation benchmarks~\cite{seo2020ytvos_rvos,khoreva2019davis_rvos,ding2023mevis}, which require segmenting objects based on text prompts. 
Although more language-oriented, these datasets are still object-centric and do not involve predicting a high-level description of the entire video.

Some approaches~\cite{ding2023mevis,yu2023convolutions,he2024dshmp} for these tasks use text and image backbones to encode the text prompt and video frames, followed by a transformer decoder and segmentation head to generate masks. 
Meanwhile, recent LLM-based approaches~\cite{yan2024visa,bai2024onetoken} use a special \texttt{SEG} vocabulary token in conjunction with a segmentation network to predict the target masks. These approaches typically utilize several image-level and video segmentation datasets for training, and utilize other models~\cite{bekuzarov2023xmem2,li2025llamavid} for dynamic frame selection or temporal mask propagation.

In contrast to these two categories of works, we propose a unified dataset that provides the annotations as well as benchmark tasks to evaluate both holistic/high-level, and pixel-level video understanding. 
Additionally, we propose a baseline architecture that is end-to-end trainable, and can be effectively trained for segmentation using only our dataset, thereby making it easy to setup and extend.

\section{\ourDataset{} Dataset}
\label{sec:our_dataset}

Our dataset is designed to evaluate both holistic as well as pixel-level video understanding. To this end, we annotate \totalVideosNew{} videos with detailed captions in which words/phrases referencing salient objects are grounded with temporally consistent segmentation masks. All annotations are done by professional human annotators.

\subsection{Video Source}
\label{subsec:video_source}

For our objective, it is crucial to annotate videos which contain meaningful events that can only be explained through effective temporal reasoning. At the same time, to effectively evaluate pixel-level understanding, the videos should feature challenging scenes with multiple moving objects.

To meet these requirements, we annotate videos from three sources: (1) Oops dataset~\cite{epstein2020oops}, a collection of `fail videos' from the internet, (2) Unidentified Video Object (UVO)~\cite{wang2021uvo}, which is a video segmentation dataset, and (3) Kinetics-700~\cite{kay2017kinetics}, a popular human action recognition dataset.
These sources are suitable for our dataset since they contain in-the-wild videos with diverse objects and backgrounds. Moreover, they contain videos with multiple objects undergoing appearance and shape changes, and fast motion. This combination of attributes makes the videos challenging both in captioning and in segmentation. 

\begin{figure*}[t]
\centering
  \begin{minipage}[b]{0.31\linewidth}
    \centering
    \includegraphics[width=\linewidth,height=4.25cm]{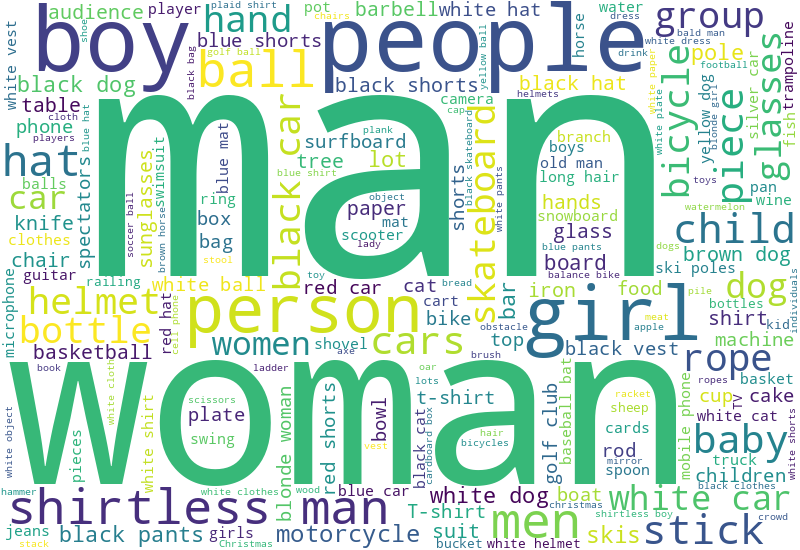}
    \subcaption{}
    \label{fig:dataset_stats_wordcloud}
  \end{minipage}%
  \hfill %
  \begin{minipage}[b]{0.33\linewidth}
    \centering
    \includegraphics[width=\linewidth,height=4.25cm]{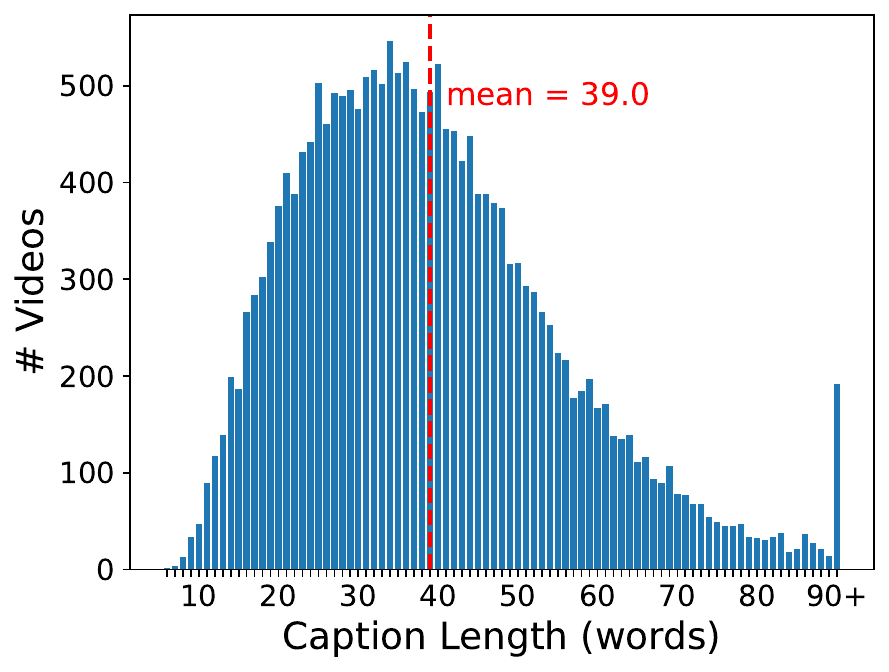} 
    \subcaption{}
    \label{fig:dataset_stats_caption_len}
  \end{minipage}%
  \hfill
  \begin{minipage}[b]{0.32\textwidth}
    \centering
    \includegraphics[width=\linewidth,height=4.25cm]{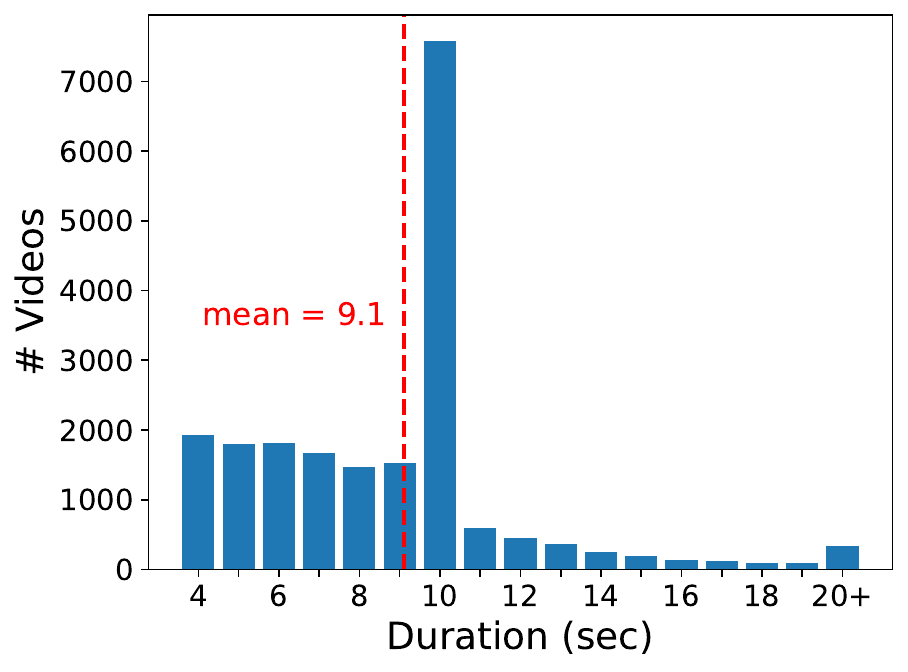} 
    \subcaption{}
    \label{fig:dataset_stats_video_duration}
  \end{minipage}
  
\caption{
\textbf{\ourDataset{} Dataset.} (a): Word cloud for nouns in grounding phrases. (b), (c): Histograms for caption length and video duration, respectively. Since these distribution are long-tailed, the right-most bar labeled `$n+$' captures all values $\geq n$ to prevent visual distortion.
}

\label{fig:dataset_stats_main}
\end{figure*}

We annotate \totalVideosNew{} videos with durations ranging from 4 to 30 seconds (distribution illustrated in Fig.~\ref{fig:dataset_stats_video_duration}). 
Note that we use only the raw videos from these Oops dataset and disregard their annotations. More details on video selection are given in supplementary.

\subsection{Annotation Process}
\label{subsec:annotation_process}

The annotation process consists of two main steps: first, a detailed caption is written (Step 1a) in which objects selected for segmentation masks are identified and marked using a specific syntax (Step 1b). In the second step, a different annotator reviews the video and its text, and draws the corresponding segmentation masks (Step 2). The process is illustrated in Fig.~\ref{fig:annotation_process}. Further details are given below.

\PAR{Step 1a: Text Captions}
Professional annotators fluent in English were tasked with writing a detailed caption covering the events in each video. To ensure consistency in the level of detail and writing style, they received comprehensive guidelines with several examples. Annotators were advised to think as if they have $\sim$30 seconds to convey the video content in as much detail as possible over an audio call. They were instructed to describe the `fail' event~\cite{epstein2020oops}, the actions and movements of objects, as well as the background scene and any additional, interesting elements. Since our dataset is video-centric, annotators were explicitly guided to include temporal information, with the instruction: ``\emph{Focus on details that require watching the entire video and cannot be inferred from a few frames.}''

\begin{figure}
    \centering
    \includegraphics[width=\linewidth]{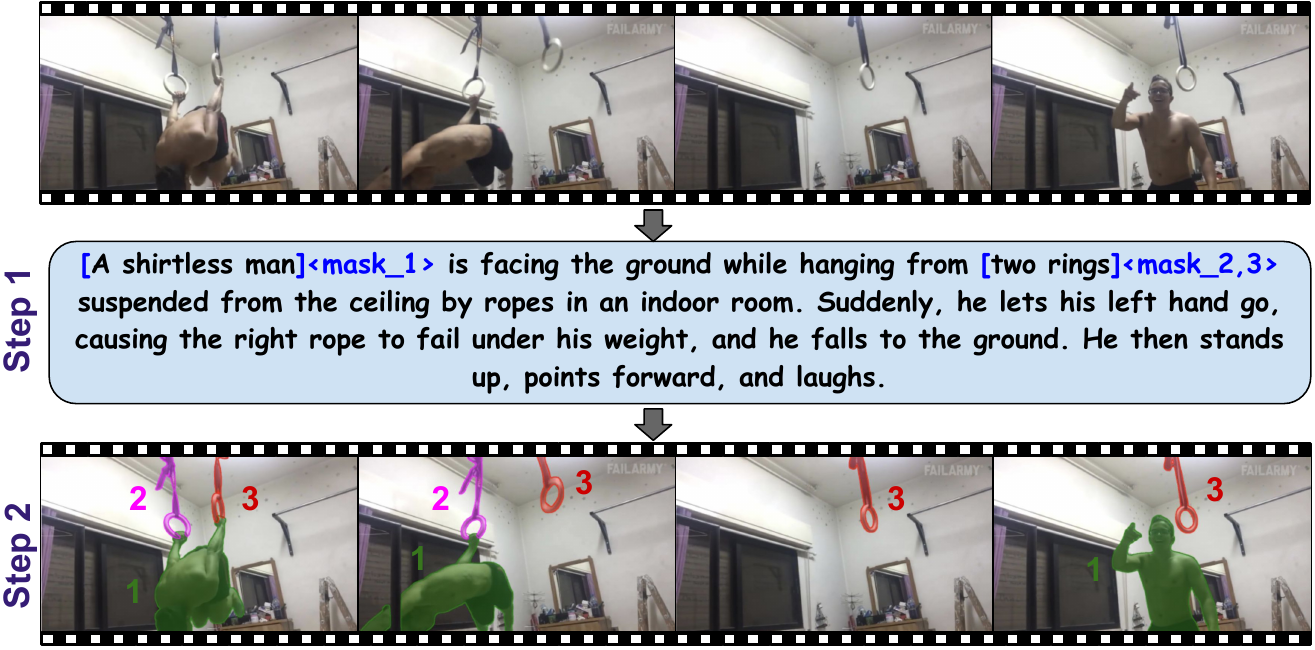}
    \caption{\textbf{Annotation Process.}
    Our annotation process consists of two steps. Step 1: A detailed video caption is written by human annotators. Salient objects are identified and marked using a special syntax (highlighted in \textcolor{blue}{blue}). Step 2: Segmentation masks are drawn for the salient objects throughout the video.
    }
    \label{fig:annotation_process}
\end{figure}

\vskip-2pt
\PAR{Step 1b: Phrase Grounding}
Alongside writing the text caption, annotators were instructed to identify salient objects/actors in the video, and mark the corresponding phrases or words using a specific syntax, as illustrated in Fig.~\ref{fig:annotation_process}. This syntax is programmatically parsable to extract object IDs that can be removed to produce a standard, human-readable caption. To streamline the mask-drawing process, annotators were advised to mark no more than 10 objects per video.

\PAR{Step 2: Segmentation Masks}
In the second step, a different annotator reviews the video and its caption, and uses a polygon tool to draw segmentation masks for each marked object.
The annotators were instructed to read the text carefully and watch the video multiple times to ensure that correct masks were drawn for the each grounding phrase. To balance annotation accuracy and cost, we annotate masks at 1 frame-per-second (fps) and then use an off-the-shelf SAM2~\cite{ravi2024sam2} model to propagate these masks to adjacent frames, producing temporally dense mask annotations at 30 fps. A similar approach has been successfully applied in other video segmentation datasets~\cite{wang2021uvo,athar2023burst}.

\PAR{Annotation Statistics and Quality Control} 
For Step 1, we employed 24 annotators and 4 quality inspectors who reviewed/corrected the text captions and language grounding. On average, each video required 9 minutes for annotation and quality checking. For Step 2, we hired 40 annotators and 5 quality inspectors, with each video frame requiring an average of 7 minutes to annotate.

\begin{table*}[t]
\centering
\small

\newcommand\RotText[2]{\rotatebox[origin=c]{90}{\parbox{#1}{\centering#2}}}
\renewcommand\tabularxcolumn[1]{m{#1}}%

\begin{tabularx}{\linewidth}{clcYcYYYYY}
    \toprule
    Type & Dataset Name                                  & Year & Human Ann. & Videos & Duration (sec) & Caption (words) & Object Classes & Object Tracks & Object Masks \\  
    \midrule
    \multirow{8}{*}{\RotText{2.4cm}{Captioning and Q/A}} & MSVD~\cite{chen2011msvd}  & 2011 & \cmarkg & 2K     &  9.7          & 8.7       &  -      &    -       & -    \\
    & MSR-VTT~\cite{xu2017_msvd_msrvtt}           & 2016 & \cmarkg & 10K           &  15.0         & 9.3         &    -          & -  &  -  \\
    & ActivityNet~\cite{yu2019activitynet}        & 2017 & \cmarkg & 20K / 100K    &  36.0         & 13.5        &    -          & -  &  -   \\
    & YouCook2~\cite{zhou2018YouCook}             & 2018 & \cmarkg & 2K / 14K      &  19.6         & 8.8         &    -          & -   &  -  \\
    & VATEX~\cite{wang2019vatex}                  & 2019 & \cmarkg & 41K           &   $\sim$10    & 15.2        &    -          & -   &  -  \\
    & DiDeMo~\cite{anne2017didemo}                & 2017 & \cmarkg & 10K / 27K     &  6.9          & 8.0         &    -          & -   &  -  \\
    & WebVid10M~\cite{bain2021webvid10m}          & 2021 & \xmarkr & 10M           &  17.2         &  14.2        &    -          & -   &  -  \\
    & Panda70M~\cite{chen2024panda70m}            & 2024 & \xmarkr & 3.8M / 70.8M  &  8.5          & 13.2         &    -          & -   &  -  \\
    \midrule
    \multirow{4}{*}{\RotText{1.5cm}{Referral VOS}} & A2D Sentence~\cite{gavrilyuk2018a2d_sent}   & 2018 & \cmarkg & 3,782  &   4.9         &   -         &  -        &     4,825     & 58K \\
    & DAVIS-17 RVOS~\cite{khoreva2019davis_rvos}  & 2019 & \cmarkg & 90            &   2.4         &  -       &  78      &     205       & 13.5K \\
    & Refer-Youtube-VOS~\cite{seo2020ytvos_rvos}  & 2020 & \cmarkg & 3,978         &   5.5         &  -       &  94      &     7,451     & 131K  \\
    & MeViS~\cite{ding2023mevis}                  & 2023 & \cmarkg & 2,006         &   13.2        &  -       &  -       &     8,171     &  443K \\
    & ReVOS~\cite{yan2024visa}                    & 2024 & \cmarkg & 1,042         &   8.6         &     -       &  -       &     9,084     &  273K \\
    \midrule
    \multirow{5}{*}{\RotText{2.1cm}{Class-guided Segmentation}} & YouTube-VIS~\cite{yang2019youtubevis} & 2019 & \cmarkg  & 2,883 & 5.6     & -       & 40       &   4,883       & 131K \\
    & OVIS~\cite{qi2022ovis}                      & 2021 & \cmarkg & 901           &  12.8         &  -     &  25     &   5,223       & 296K \\
    & UVO~\cite{wang2021uvo}                      & 2021 & \cmarkg & 11,228        &    9.6        & -      &  80    &  104,898      & 593K \\
    & BURST~\cite{athar2023burst}                 & 2023 & \cmarkg & 2,914         &   35.7        & -      & 482     &   16,089      & 600K \\
    & LV-VIS~\cite{wang2023lvvis}                 & 2023 & \cmarkg & 4,828         &    4.6        &  -     & 1,196     &   25,588      & 544K \\
    \midrule
    & \textbf{\ourDataset{} (Ours)}               & 2025 & \cmarkg & \totalVideosNew         &   9.1  &  39.0    & 11,492     &    65,588     & 18.3M \\
    \bottomrule
    
\end{tabularx}

\caption{\textbf{Dataset Comparison.} In the `Videos' column, entries formatted as `X / Y' indicate a total of X videos with Y labeled sub-clips, with the average sub-clip duration being reported.
}

\label{tab:dataset_comparison_v2}
\end{table*}

\subsection{Benchmark Design and Evaluation Tasks}
\label{subsec:benchmark_tasks}

Leveraging our annotations, we create a benchmark to evaluate both high-level video understanding and pixel-precise segmentation. Consequently, it includes two tasks:
    
\PAR{1) Video Captioning} In this task, the model is expected to produce an open-ended text summary that explains the events in the video, including descriptions of salient objects and background elements.

\PAR{2) Language-Guided Video Instance Segmentation (LG-VIS)} For this task, the model receives a text prompt describing a specific set of objects, \eg, `\emph{Where is the person who walks from left to right?}', and is expected to predict segmentation masks for the corresponding objects. For cases where multiple objects are referenced, \eg,  `\emph{Where is the group of kittens that are playing?}', a separate mask is required for each object. This is in contrast to Referral Video Object Segmentation~\cite{seo2020ytvos_rvos,ding2023mevis,khoreva2019davis_rvos}, where only a single, binary mask covering all target objects is required.

From a vision-language perspective, both tasks are forms of Video Question-Answering. For Video Captioning, the question asks for a video description, with the answer provided as open-ended text. For LG-VIS, the question is a `\emph{Where is?}' query regarding specific objects, and the answer is a video-length segmentation mask for each object.

\PAR{Converting Grounded Captions to LG-VIS Prompts} As outlined in Sec.~\ref{subsec:annotation_process}, our annotations contain phrase grounding for labeled objects. We use GPT-4~\cite{achiam2023gpt4} to convert these grounded captions into `\emph{Where is?}' style questions, which serve as text prompts for the LG-VIS task. Further details about this process are provided in supplementary.

\PAR{Dataset Splits}
We split the \totalVideosNew{} videos into train, validation, and test sets with 14,516 videos for training, and 2,950 each for validation and testing. Statistics for each set are given in supplementary.

\subsection{Comparison with Related Datasets}
\label{subsec:dataset_comparison}

Table~\ref{tab:dataset_comparison_v2} provides a quantitative comparison between \ourDataset{} and existing datasets, showing that \ourDataset{} uniquely combines text captions and pixel-precise segmentation masks.

\PAR{Captioning and Q/A} For text descriptions, \ourDataset{} prioritizes quality over quantity, offering detailed, high-quality, human-written captions with phrase grounding for objects. 
Our captions average 39 words (distribution illustrated in Fig.~\ref{fig:dataset_stats_caption_len}), significantly longer than those in other datasets. 
While \ourDataset{} annotates more videos than early datasets like MSVD~\cite{chen2011msvd} and MSR-VTT~\cite{xu2016msrvtt}, later datasets with temporally dense captions, such DiDeMo~\cite{anne2017didemo}, and YouCook2~\cite{zhou2018YouCook}, include more labeled video segments. 
Recently, large-scale datasets such as WebVid10M~\cite{bain2021webvid10m} and Panda70M~\cite{chen2024panda70m} have emerged, containing millions of \textit{automatically} captioned videos sourced from the internet. 
While useful for (pre-)training large models, these datasets offer shorter, less detailed descriptions and have neither segmentation masks nor phrase grounding.

\PAR{Referral VOS} Compared to Referral Video Object Segmentation (VOS) datasets, \ourDataset{} stands out by providing holistic, video-level captions in addition to object-level language expressions. 
With \totalVideosNew{} annotated videos, \ourDataset{} is significantly larger than the second-largest Refer-YouTubeVOS~\cite{seo2020ytvos_rvos}, which contains only 3,978 videos. 
Although MeViS~\cite{ding2023mevis} features longer videos, averaging 13.2s compared to our 9.1s, \ourDataset{} offers significantly more annotated objects (65,588) compared to MeViS (8,171).

\PAR{Class-Guided Segmentation} Compared to these datasets, \ourDataset{} provides more nuanced language labels, going beyond single-word category labels typical in Video Instance Segmentation (VIS)~\cite{yang2019youtubevis,qi2022ovis} datasets. 
Using NLTK~\cite{bird2009NLTK}, we found that our object-grounding phrases contains 11,492 unique nouns/noun-phrases (illustrated in Fig.~\ref{fig:dataset_stats_wordcloud}), which is significantly higher than the 1,196 object categories in LV-VIS~\cite{wang2023lvvis}.
Furthermore, \ourDataset{} contains more videos than the others, and also contains more labeled object tracks than all other datasets except UVO~\cite{wang2021uvo}.
Lastly, \ourDataset{} stands out by providing detailed text captions in addition to the segmentation masks.

\section{Evaluation Measures}
\label{sec:eval_measures}

\vspace{-8pt}

\begin{table}[ht]
\centering
\footnotesize

\newcommand\RotText[2]{\rotatebox[origin=c]{90}{\parbox{#1}{\centering#2}}}
\newcommand\NonRotText[1]{\parbox[c]{\linewidth}{\centering#1}}
\renewcommand\tabularxcolumn[1]{m{#1}}%

\begin{tabularx}{\linewidth}{p{0.08cm}lp{0.02cm}YYYp{0.02cm}Y}
    \toprule
     & \multirow{2}{*}{Evaluation Method}               & & \multicolumn{3}{c}{Part 1} & \multicolumn{2}{r}{Part 2} \\
     \cmidrule{4-6}\cmidrule{8-8}
     & & & {\small $r$}  & {\small $r_\text{max}$} & {\small $r_\text{avg}$ } & & $\Delta_\text{PN}$ \\  
    \cmidrule{1-2}\cmidrule{4-6}\cmidrule{8-8}
    \multirow{3}{*}{\RotText{1cm}{LLM}}  & GPT4~\cite{achiam2023gpt4}                        & & 58.9   &   62.1   &  \textbf{67.3}   & & \textbf{37.6}  \\
    & Llama3-70B~\cite{dubey2024llama3} & &  \textbf{59.3}   &   \textbf{64.3}   &  65.6   & & 29.2 \\
    & Llama3-8B~\cite{dubey2024llama3}                  & & 31.3   &   -      &  45.9   & & 2.4 \\
    \midrule
    \multirow{6}{*}{\RotText{1.9cm}{Embedding Sim.}}  & AnglE Lm2-7B~\cite{li2023angle} & & 50.5   &   50.0   &  50.3   & & 3.3  \\ 
    & STSB Roberta-L~\cite{Reimers2019-sentencebert}      & & 49.5   &   48.3   &  49.1   & & 7.5  \\ 
    & AnglE Lm2-13B~\cite{li2023angle}                    & & 45.6   &   47.4   &  46.4   & & 4.5  \\ 
    & AnglE UAE-L~\cite{li2023angle}                      & & 44.5   &   42.7   &  40.5   & & 0.9  \\ 
    & MPNet-B~\cite{Reimers2019-sentencebert}             & & 41.5   &   35.4   &  34.7   & & 0.4 \\
    & BERTScore~\cite{zhang2020bertscore}                 & & 22.1   &   22.9   &  26.2   & & -2.4 \\
    \midrule
    \multirow{4}{*}{\RotText{1.2cm}{Word mat.}}  & METEOR~\cite{banerjee2005meteor} & & 22.3   &   23.9   &  22.8   & & -16.4 \\
    & BLEU4~\cite{papineni2002bleu}                       & & 19.9   &   15.9   &  15.4   & & -27.6 \\
    & ROUGE-L~\cite{lin2004rouge}                         & & 18.1   &   18.2   &  13.2   & & -23.7 \\
    & CIDEr~\cite{vedantam2015cider}                      & & 2.8    &   0.1    &  -0.8   & & -21.2 \\
    \bottomrule
    
\end{tabularx}

\caption{
\textbf{Caption Scoring User Study.} Results for Part 1 (Correlation with Human Scores) and Part 2 (Robustness to Hard Positives/Negatives). 
$r$, $r_{max}$ and $r_{avg}$ denote Pearson correlation coefficients. $\Delta_{PN}$ is the absolute difference in predicted scores.
}

\label{tab:user_study_eval_metrics}
\end{table}

As outlined in Sec.~\ref{subsec:benchmark_tasks}, our benchmark comprises two tasks: Video Captioning and LG-VIS. 
Evaluating the video caption quality is challenging since it requires computing the similarity between two open-ended text passages—the human-written ground truth and the model’s prediction. To address this, we first conduct a comprehensive user study for scoring open-ended text similarity in Sec.~\ref{sec:user_study}, before introducing the selected evaluation measures in Sec.~\ref{sec:selected_metrics}.

\subsection{User Study for Video Caption Scoring}
\label{sec:user_study}

To decide evaluation measures for video captioning, we carried out a two-part user study comparing several scoring methods which are categorized into three groups: (1) classical text similarity metrics, \eg, METEOR~\cite{banerjee2005meteor} and BLEU4~\cite{papineni2002bleu}, which rely on word/phrase matching; (2) embedding similarity-based measures, \eg, BERTScore~\cite{zhang2020bertscore} and AnglE~\cite{li2023angle}, which summarize the ground-truth and prediction as an embedding using a language model, followed by computing the cosine similarity between the two; and (3) recent LLM-based scoring methods~\cite{maaz2023videochatgpt}, where an LLM is provided with the ground-truth and predicted captions along with instructions for assessing similarity, and it outputs a similarity score directly as text.

\PAR{Part 1: Correlation with Human Scores} In this part, 20 participants rated the accuracy of various model-predicted video captions on a scale of 0 to 10, with a minimum interval of 0.5. Similar to the evaluation methods, participants could only read the ground-truth caption and could not watch the video itself. They represented 10 different nationalities and had either native, or professional English proficiency. A total of 131 video captions were evaluated, with each sample scored by two different individuals.

To filter samples with diverging human scores, we calculated the difference $\Delta$ between the two assigned scores for each sample, discarding any with $\Delta > \alpha$ (\ie, outlier removal). We used multiple thresholds $\{\alpha_i\}$ ranging from 0.5 to 2.5 in steps of 0.5. For each $\alpha_i$ and each method, we calculated the Pearson correlation coefficient $r_i$ between averaged human scores and the predicted scores. The final correlation score $r$ for each method was computed as the mean of the set $\{r_i\}$, and is reported in Table~\ref{tab:user_study_eval_metrics}.

Results indicate that LLM-based scoring methods performed best, with Llama3-70B~\cite{dubey2024llama3} achieving $r=59.3\%$, closely followed by GPT-4~\cite{achiam2023gpt4} with 58.9\%. Model size significantly impacted performance, as evidenced by Llama3-8B's lower $r=31.3\%$. Among embedding-based methods, AnglE Llama2-7B~\cite{li2023angle} performed best at 50.5\%, while classical word/phrase matching metrics lagged behind, with METEOR~\cite{banerjee2005meteor} achieving only $r=22.3\%$.

\PAR{Pooling Scores Across Multiple Ground Truths} A common strategy to enhance robustness in text similarity evaluation is to compare predictions with multiple, synonymous ground-truths. Using GPT-4~\cite{achiam2023gpt4}, we generated four rephrasings for each human-written caption, resulting in five ground-truth variants per video. For each video, we calculated the similarity between the prediction and all ground-truths, followed by applying either average or maximum pooling to obtain the final score. We report the resulting correlations as $r_\text{avg}$ and $r_\text{max}$ in Table~\ref{tab:user_study_eval_metrics}. 
As shown in the table, LLM-based scoring methods benefited most from average pooling, with Llama3-70B improving from 59.3\% to 65.6\% and GPT-4 from 58.9\% to 67.3\%. Embedding-based methods generally saw minimal improvements for both types of pooling, with some, like MPNet-B and AnglE-UAE-L, experiencing slight declines. Word/phrase matching metrics similarly showed no substantial improvement.

\begin{figure*}[h]
\centering
\includegraphics[width=\linewidth]{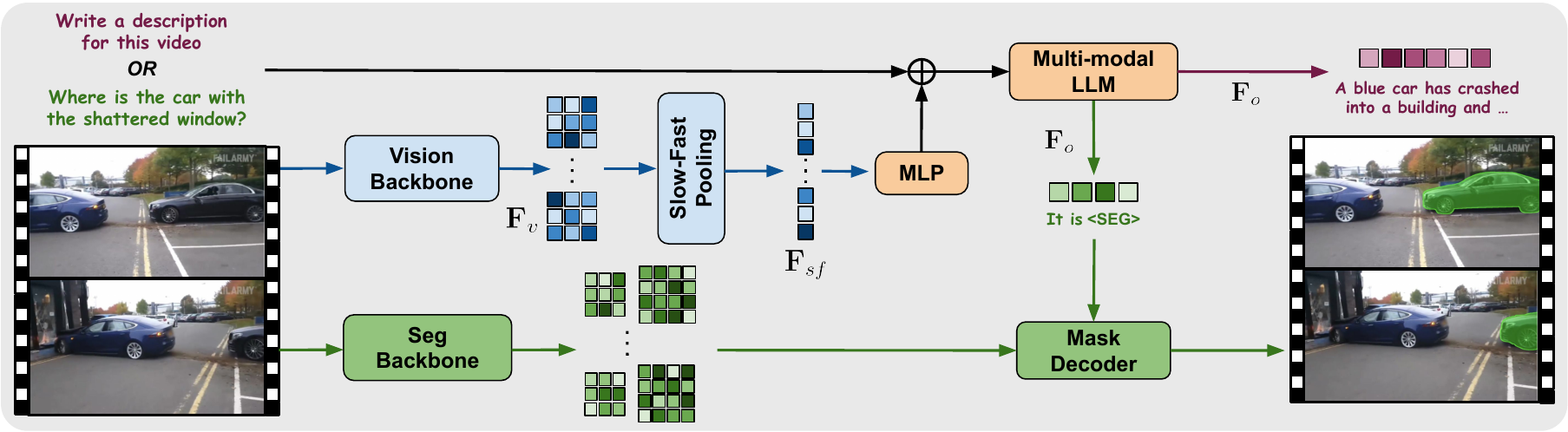}
\caption{
\textbf{\ourBaseline{} Architecture.} The vision backbone and projection MLP encode the input video frames into a set of features $\mathcal{F}_{sf}$ which are concatenated with the text embeddings and input to the LLM. For \textcolor[HTML]{741b47}{Video Captioning}, the output $\mathbf{F}_o$ is decoded into text. For \textcolor[HTML]{38761d} {LG-VIS}, the $\texttt{<SEG>}$ token in $\mathbf{F}_o$ is applied to the mask decoder along with multi-scale features from the segmentation backbone.
}
\label{fig:network_arch}
\end{figure*}

\PAR{Part 2: Robustness to Hard Positives/Negatives} Here, participants watched videos from our dataset and read the ground-truth captions (GT), and were then asked to write a hard-positive (HP) and a hard-negative (HN) caption for each video. For the HP, they rephrased the caption using different wording and style while retaining all original information. For the HN, they altered critical details while retaining most wording and sentence structure. Each of the 10 participants processed 5 videos, yielding 50 samples.

We computed the similarity between GT-HP and GT-HN pairs for each video using different methods. 
The scores were scaled to $[0,100]$\footnote{Except CIDEr, which is not a normalized metric} and the difference $\Delta_\text{PN}$ between the GT-HP and GT-HN pairs was calculated, and reported in Table~\ref{tab:user_study_eval_metrics}. 
Ideally, $\Delta_\text{PN}$ should be high, indicating that methods assign higher similarity to GT-HP pairs and lower similarity to GT-HN pairs. 
GPT-4 achieved the highest $\Delta_\text{PN}$ (37.6), followed by Llama3-70B (29.2). 
Most embedding-based models showed low $\Delta_\text{PN}$, indicating similar scores for GT-HP and GT-HN pairs, while word/phrase matching metrics performed worst, with negative $\Delta_\text{PN}$ values suggesting higher scores for GT-HN pairs than GT-HP pairs.

\vskip-4pt
\subsection{Selected Evaluation Measures} 
\label{sec:selected_metrics}

\vspace{-4pt}

\vskip-4pt
\PAR{Video Captioning} Although GPT-4 achieved the highest $r_\text{avg}$ and $\Delta_\text{PN}$ in our study, it is proprietary and subject to updates, which complicates reproducibility. Moreover, its runtime can be slow due to token limits, and it incurs monetary costs which effectively makes the benchmark `pay-to-evaluate'. To support open-source, reproducible research, we accept a slight performance trade-off and select Llama3-70B as the evaluation measure for our video captioning task. Since $r_\text{avg}$ is noticeably higher than $r$ for Llama3-70B (Table~\ref{tab:user_study_eval_metrics}), we generate 4 reworded variants of each ground-truth caption, and compute the final accuracy by averaging over the prediction-GT scores for each video. This score lies in $[0,5]$ and is abbreviated as CA (Caption Accuracy).

\PAR{Language-Guided Video Instance Segmentation (LG-VIS)} Unlike text similarity, segmentation mask accuracy is a well-studied problem with several feasible metrics~\cite{yang2019youtubevis,luiten2020HOTA,Li2022teta}. We select Track mean Average Precision (mAP) as our primary metric since it is widely used by video segmentation benchmarks~\cite{yang2019youtubevis,qi2022ovis,wang2021uvo} and accommodates multi-object prediction.

\vspace{-5pt}

\section{\ourBaseline{} Model}
\label{sec:baseline_model}

We propose an effective baseline called \ourBaseline{} that can tackle both Video Captioning and LG-VIS with a single, end-to-end trained model.

\subsection{Architecture}
\label{subsec:network_architecture}

\ourBaseline{} extends the popular LLaVA~\cite{liu2023llava,liu2024llavanext,liu2023improvedllava} architecture with segmentation ability. It is illustrated in Fig.~\ref{fig:network_arch} and comprises 3 main parts: (1) a multi-modal LLM, (2), a vision backbone, and (3) a segmentation network.

\PAR{Vision Features} We input $T$ uniformly sampled frames from the video to the vision backbone, yielding a set of video features $\mathbf{F}_v \in \mathbb{R}^{T\times H\times W\times C}$. Next, we adopt the workflow proposed by Xu~\etal~\cite{xu2024sfllava} and split $\mathbf{F}_v$ into two sets of features, namely `slow' features $\mathbf{F}_v^s \in \mathbb{R}^{T_s \times H\times W\times C}$ and `fast features' $\mathbf{F}_v^f \in \mathbb{R}^{T\times H_f \times W_f\times C}$. Fast features are generated by aggressively downsampling the spatial dimensions in $\mathbf{F}_v$ using adaptive average pooling, \ie $H_f << H$ and $W_f << W$. Meanwhile, slow features are obtained by uniformly sampling $T_s < T$ frame features from $\mathbf{F}_v$. These two sets of features are then flattened and concatenated to yield the set of `slow-fast' video features $\mathbf{F}_{sf} \in \mathbb{R}^{N_v\times C}$ where $N_v = \left(T\times H_f\times W_f\right) + \left( T_s\times H\times W \right)$. The idea is to retain finegrained information along both temporal and spatial dimensions: the slow features are temporally dense but spatially condensed, whereas the fast features are spatially dense but temporally sparse. This reduces the number of video tokens $N_v$ input to the multimodal LLM.

\PAR{Multi-modal LLM} The slow-fast video features are then applied to a projection MLP, concatenated with the text prompt embeddings, and then input to the LLM, which outputs a set of embeddings $\mathbf{F}_o$. For the Video Captioning task, $\mathbf{F}_o$ is decoded into text. For LG-VIS, the model is trained to output a special \texttt{<SEG>} token which is used by the segmentation network to regress the mask. Note that when the text prompt references multiple objects, the model generates a separate \texttt{<SEG>} token for each target object.

\vspace{-5pt}

\PAR{Segmentation Network} The video frames are encoded at high resolution by a segmentation backbone to yield multi-scale features which, together with \texttt{<SEG>} tokens, are input to a mask decoder to obtain the final segmentation masks. The mask decoder architecture is borrowed from SAM2~\cite{ravi2024sam2} and consists of multiple layers of bi-directional cross-attention between the \texttt{<SEG>} tokens and the segmentation backbone features. Finally, the dot-product between the two is computed to obtain the mask logits. We refer to Ravi~\etal~\cite{ravi2024sam2} for more details. Our approach for mask prediction using \texttt{<SEG>} tokens generated by an MLLM is inspired from recent image segmentation~\cite{lai2024lisa,yuan2024osprey,rasheed2024glamm,yang2023lisa++} and video segmentation works~\cite{yan2024visa,bai2024onetoken}.

\PAR{Comparison to Existing Methods}
\ourBaseline{} is a compact baseline model that is easy to setup and extend, unlike existing LLM-based video segmentation approaches~\cite{yan2024visa,bai2024onetoken}: VISA~\cite{yan2024visa} involves inference with a pretrained Llama-Vid~\cite{li2025llamavid} to select key frames for their own model, and both VISA and VideoLISA~\cite{bai2024onetoken} rely on pretrained models~\cite{cheng2022xmem,bekuzarov2023xmem2} to perform/improve temporal mask propagation. Moreover, these approaches tackle only segmentation tasks whereas our model is trained to tackle both captioning and segmentation.

\subsection{Implementation Details}
\label{subsec:implementation_details}

We employ an AM-RADIO~\cite{ranzinger2024radio} ViT-H~\cite{beyer2021vit} model as the vision backbone and Llama3-8B~\cite{dubey2024llama3} as the LLM. The segmentation backbone is a Hiera-Small~\cite{ryali2023hiera} network which, together with the mask decoder, is initialized with weights from SAM2~\cite{ravi2024sam2}. We sample $T=32$ frames and use $T_s=8$ slow frames. We train our model in three stages: (1) a vision-language alignment stage to optimize the projection MLP, (2) a finetuning step to optimize the MLLM, projection MLP and vision backbone for video captioning, and finally (3) the entire model with the segmentation network is fintuned for both tasks. For the first two stages, we utilize a total of 3.5M video caption samples from WebVid10M~\cite{bain2021webvid10m} and Panda70M~\cite{chen2024panda70m}. In stage 3, we finetune the model on the \ourDataset{} training set for both tasks. Training takes a total of $\sim4$ days on 32 A100 GPUs. Further implementation details can be found in supplementary.

\subsection{Ablation Studies}
\label{subsec:ablations}

We report ablations in Table~\ref{tab:ablations} and discuss them below. %

\begin{table}[h]
\centering
\small

\newcommand\RotText[2]{\rotatebox[origin=c]{90}{\parbox{#1}{\centering#2}}}
\newcommand\NonRotText[1]{\parbox[c]{\linewidth}{\centering#1}}
\renewcommand\tabularxcolumn[1]{m{#1}} %
\newcommand\APat[1]{$\text{AP}_\text{#1}$}

\begin{tabularx}{\linewidth}{p{0.1cm}ccccccc}
    \toprule
    \#             & Vis Enc. & $T_s$   & SF-Pool.    & Cap.    & Seg. & CA  &  mAP  \\
    \midrule
    1              & AM-RAD.  &   1     & \xmarkr & \cmarkg & \cmarkg  & 2.1 & 15.9  \\
    2              & AM-RAD.  &   2     & \xmarkr & \cmarkg & \cmarkg  & 2.4 & 17.0  \\
    3              & AM-RAD.  &   4     & \xmarkr & \cmarkg & \cmarkg  & 2.8 & 18.8  \\
    4              & AM-RAD.  &   8     & \cmarkg & \cmarkg & \xmarkr & 3.0 & -     \\
    5              & AM-RAD.  &   8     & \cmarkg & \xmarkr & \cmarkg & -   & 18.5  \\
    6              & CLIP     &   8     & \xmarkr & \cmarkg & \cmarkg & 2.9 & 19.9  \\
    \midrule
                  & AM-RAD. &  8      & \cmarkg & \cmarkg & \cmarkg & 3.0 & 20.5 \\
    \bottomrule
    
\end{tabularx}

\caption{Ablation experiments on the ViCaS validation set. CA: Caption Accuracy. SF-Pool: Slow-Fast Pooling. 
}

\label{tab:ablations}
\end{table}

\vspace{-5pt}

\PAR{Input Frames}
In rows 1-3, we train our model using $T_s= \{ 1,2,4 \}$ frames instead of $Ts=8$ in the final setting. We see that using fewer frames strongly reduces performance, especially for captioning, showing that our dataset/benchmark is video-centric and requires temporal context to effectively tackle.

\PAR{Task Synergy}
For rows 4 and 5, we train the model for either one of our benchmark tasks. We see that the model trained only for LG-VIS achieves 18.5 mAP which is much worse than the 20.5 achieved by the multi-task model. Meanwhile, the Caption Accuracy remains unchanged at 3.0.
This indicates that pixel-level segmentation benefits greatly from holistic video understanding.

\PAR{LLaVA Baseline}
Row 6 shows results for a LLaVA-NeXT baseline trained with our recipe which uses a CLIP vision encoder and no Slow-Fast Pooling. This achieves 19.9 mAP and 2.9 CA, both of which are worse than our final setting.

\subsection{Benchmark Results}
\label{subsec:benchmark_results}

\vspace{-5pt}

\begin{table}[h]
\centering
\footnotesize

\newcommand\RotText[2]{\rotatebox[origin=c]{90}{\parbox{#1}{\centering#2}}}
\newcommand\NonRotText[1]{\parbox[c]{\linewidth}{\centering#1}}
\renewcommand\tabularxcolumn[1]{m{#1}} %
\newcommand\APat[1]{$\text{AP}_\text{#1}$}

\begin{tabularx}{\linewidth}{lYYYYY}
    \toprule
    Model                               & CA & mAP   & \APat{50} & \APat{75} & \APat{90} \\
    \midrule
    LLaVA-OV~\cite{li2024llavaonevision} (ZS)                    &  2.9   & - & - & - & - \\
    MiniCPM-o 2.6 (ZS)                                           & 3.0    & - & - & - & - \\
    LMPM~\cite{ding2023mevis}          & -                                &  8.4 &  17.1     &   7.7    &    1.8    \\
    DsHmp~\cite{he2024dshmp}           &   -                             & 13.3 &  27.6     &   11.6    &    2.9    \\
    VideoLISA~\cite{bai2024onetoken}   &   - & 10.7 & 23.8 & 8.5 & 1.8 \\
    \midrule
    \ourBaseline{}                     &  3.0      &  20.5 &  38.6     &   19.0    &    5.1    \\
    \bottomrule
    
\end{tabularx}

\vspace{-4pt}
\caption{Benchmark Results on our validation set. Refer to supplementary for test set results. CA: Caption Accuracy}

\label{tab:benchmark_results}
\end{table}

\vspace{-5pt}

Table~\ref{tab:benchmark_results} compares \ourBaseline{} to other methods on the validation set. For video captioning, we evaluate off-the-shelf LLaVA-OneVision~\cite{li2024llavaonevision} and MiniCPM-o\cite{yao2024minicpm} models which achieve 2.9 and 3.0 CA scores, respectively. This is the same as \ourBaseline{}, but these models are trained on significantly more data and cannot tackle LG-VIS. For LG-VIS, we evaluate multiple existing Referral-VOS approaches. LMPM is an earlier transformer-based approach which only achieves 8.4 mAP. DsHmp~\cite{he2024dshmp} performs better with 13.3 mAP since the architecture decouples static and motion cues to improve video segmentation performance. Finally, VideoLISA~\cite{bai2024onetoken} is a recent LLM-based approach which only achieves 10.7 mAP when finetuned on ViCaS. By contrast, \ourBaseline{} achieves 20.5 mAP while tackling both captioning and LG-VIS. Furthermore, unlike existing Referral-VOS approaches, it can predict multiple segmentation masks for a single prompt.

\vspace{-4pt}
\section{Conclusion}
\label{sec:conclusion}

\vspace{-3pt}

We introduce \ourDataset{}, a first-of-its-kind, human-annotated dataset that provides detailed captions for videos along with phrase-grounded segmentation masks for salient objects. Our associated benchmark comprises two tasks: (1) Video Captioning, which evaluates high-level, holistic understanding, and (2) Language-Guided Video Instance Segmentation (LG-VIS), which evaluates pixel-level understanding. We propose evaluation measures for open-ended caption accuracy which are based on open-source models, and are experimentally verified through a comprehensive user study. Furthermore, we propose \ourBaseline{}, a compact baseline which effectively tackles both tasks. 

\PAR{Acknowledgements}
We thank the participants of the user study and all the annotation personnel for their effort.

{
    \small
    \bibliographystyle{ieeenat_fullname}
    \bibliography{main}
}

\clearpage
\setcounter{section}{0}

\renewcommand{\thesection}{\Alph{section}}

\maketitlesupplementary

\section{Implementation Details}
\label{sec:supp_impl_details}

We provide the full set of implementation details and training hyperparameters for our \ourBaseline{} model here.

\PAR{Vision Backbone} Our vision backbone is a pretrained AM-RADIO~\cite{ranzinger2024radio} ViT-H/16 model. The video frames are resized by scaling the longer dimension to $384$ in aspect-ratio preserving manner, followed by zero padding to obtain a square image with size $384\times 384$. The vision backbone has a stride of 16, thus yielding $24\times 24=576$ tokens per frame. We sample a total of $T=32$ frames per video. Of these $T_s=8$ are encoded as slow frames. Meanwhile, all $T=32$ fast frames are resized to $H_f=W_f=4$ using adaptive average pooling, following the original work by Xu~\etal~\cite{xu2024sfllava}. Thus, each `fast' frame is encoded using $4\times4=16$ tokens.  Overall, the input video is represented as $N_v=(32\times 16)+(8\times 576)=5120$ tokens at the input to the LLM.

\PAR{Segmentation Network} The segmentation backbone is a Hiera-Small~\cite{ryali2023hiera} with a Feature Pyramid Network (FPN)~\cite{lin2017feature}. Since finegrained details are needed to predice accurate segmentation masks, we use a larger input resolution of $1024\times 1024$ for the segmentation network, following the original implementation from Ravi~\etal~\cite{ravi2024sam2}. The backbone Hiera model has a stride of 16, thus resulting in feature maps of size $64\times 64$. The FPN yields two high-resolution feature maps at $8\times$ and $4\times$ strides, \ie $128\times 128$ and $256\times 256$, respectively. These feature maps are used in the final stages of the mask decoder~\cite{wang2021max} to predict high-resolution segmentation masks. The entire segmentation network (backbone, FPN, and mask decoder) is initialized with pretrained weights from SAM2~\cite{ravi2024sam2}.

\PAR{Training} Our \ourBaseline{} model is trained in three stages:

\begin{itemize}
    \item Stage 1: Pretraining stage where only the projection MLP is optimized for video captioning in order to align vision and language features. 
    \item Stage 2: The projection MLP, vision backbone and LLM are optimized for video captioning.
    \item Stage 3: The entire model (projection MLP, vision backbone, LLM, and segmentation network) are optimized
\end{itemize}

Details about each stage of training are given in Table~\ref{tab:supp_train_hyperparams}. Note that we only use a small fraction of the data from WebVid10M~\cite{bain2021webvid10m} and Panda70M~\cite{chen2024panda70m}: for stage 1 we utilize 750,000 samples from each dataset, and for stage 2 we utilize 1,000,000 samples from each.

\section{Grounded Captions to LG-VIS Prompts}

\begin{figure}[h!]
    \centering
    \includegraphics[width=\linewidth]{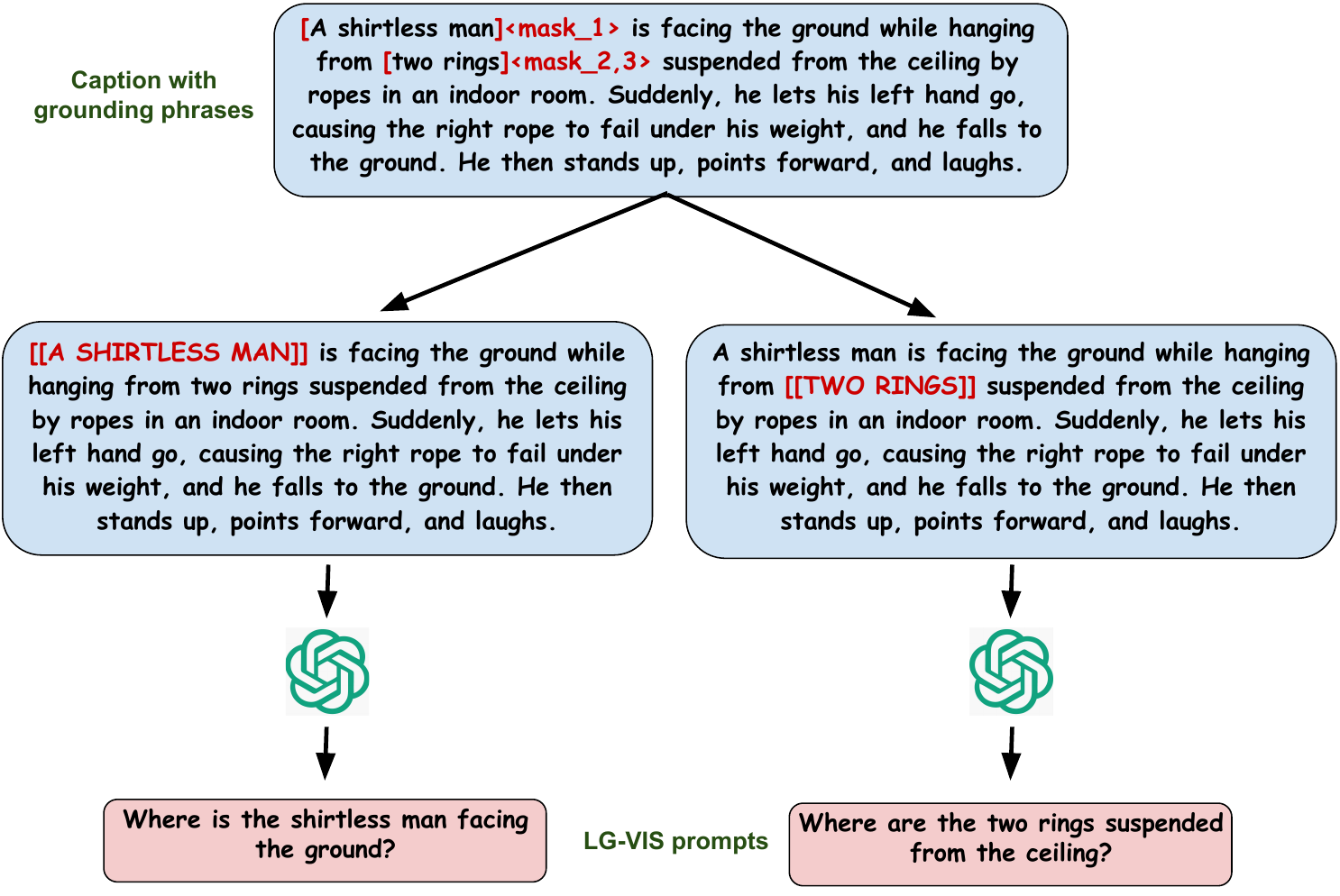}
    \caption{\textbf{Grounded captions to LG-VIS prompts.} We use GPT4~\cite{achiam2023gpt4} to convert our human-written captions with phrase grounding to phrase-specific prompts for LG-VIS. Note that the text highlighted in \textcolor{red}{red} is for ease of visualization only.}
    \label{fig:supp_cap_to_lgvis}
\end{figure}

As mentioned in Sec.~3.3 of the main text, our benchmark comprises a Language-Guided Video Instance Segmentation (LG-VIS) task which requires segmenting multiple objects based on a language prompt. To obtain these prompts from our grounded captions, we use a simple pipeline which is illustrated in Fig.~\ref{fig:supp_cap_to_lgvis}. We input each grounding phrase to GPT4 with the phrase itself being highlighted using a special syntax as shown in the second row of the figure. GPT4 is given instructions to generate a `\emph{Where is?}' style question that references the object(s) in the grounding phrase. It is told to avoid putting too much information in the prompt and only include just enough information for the prompt to be unambiguous. The output is an LG-VIS prompt for each grounding phrase.

\section{Dataset Statistics}

\begin{table*}[t!]

\newcommand\RotText[2]{\rotatebox[origin=c]{90}{\parbox{#1}{\centering#2}}}
\newcommand\NonRotText[1]{\parbox[c]{\linewidth}{\centering#1}}
\renewcommand\tabularxcolumn[1]{m{#1}} %
\newcommand{\expnumber}[2]{{#1}\mathrm{e}{#2}}
\footnotesize

\centering
\begin{tabularx}{\linewidth}{lYYY}

\toprule
Stage         &      1      &        2         &       3          \\
\midrule 
Tasks         &   VC        &   VC             &   VC, LG-VIS     \\
Datasets      &  WebVid10M~\cite{bain2021webvid10m}, Panda70M~\cite{chen2024panda70m}  & WebVid10M~\cite{bain2021webvid10m}, Panda70M~\cite{chen2024panda70m}  & ViCaS, MeViS~\cite{ding2023mevis}, Ref-YTVOS~\cite{seo2020ytvos_rvos} \\
Epochs        &    1        &       1          &    10            \\
Iterations    &  5,860      &     15,625       &    10,752        \\
Batch Size    &   256       &      128         &    128            \\       
Optimized Components & Projection MLP & Projection MLP + Vision Backbone + LLM & Projection MLP + Vision Backbone + LLM + Segmentation Network \\
Learning Rate &  1e-3  &  Vision backbone: 1e-6, Rest: 2e-5 & Vision backbone: 1e-6, Rest: 2e-5 \\
\bottomrule
    
\end{tabularx}

\caption{
\textbf{Training details for various stages.} VC: Video Captioning. LG-VIS: Language-Guided Video Instance Segmentations.}
\label{tab:supp_train_hyperparams}
\end{table*}

\begin{table*}[t!]
\centering
\footnotesize

\newcommand\RotText[2]{\rotatebox[origin=c]{90}{\parbox{#1}{\centering#2}}} 
\newcommand\NonRotText[1]{\parbox[c]{\linewidth}{\centering#1}}
\renewcommand\tabularxcolumn[1]{m{#1}} %
\newcommand\APat[1]{$\text{AP}_\text{#1}$}

\begin{tabularx}{\linewidth}{lYYYYYYY}
    \toprule
    Split      & Videos & Avg Duration (seconds) & Avg Caption (words) & Object Tracks & LG-VIS Prompts & Object Masks (Human) & Object Masks (Automatic) \\
    \midrule
    Train      & 14,516 & 9.0              & 38.8                &  46,235       & 42,024         & 445,368              & 12.3M                    \\
    Validation & 2,950  & 8.7              & 38.2                &  9,265        & 8,393          & 87,054               & 2.4M                    \\
    Test       & 2,950  & 9.8              & 40.6                &  10,088       & 9,019          & 104,661              & 2.9M                    \\
    All        & 20,416 & 9.1              & 39.0                & 65,588        & 59,436        & 637,083               & 17.7M                     \\
    \bottomrule
    
\end{tabularx}

\caption{
\textbf{Dataset statistics for train, validation and test splits.} As mentioned in Sec.~3.2, professional human annotators draw segmentation masks at 1fps, followed by using an off-the-shelf SAM2~\cite{ravi2024sam2} model to increase the temporal density to 30 fps. Both types of mask annotations are provided separately in the last two columns.
}

\label{tab:supp_split_stats}
\end{table*}

The statistics for the train, validation and test sets of our \ourDataset{} dataset are given in Table~\ref{tab:supp_split_stats}. Note that our test set has slightly higher object density than the train and validation sets, presenting a more challenging evaluation scenario.

\section{Benchmark Results (Test Set)}

\begin{table}[h]
\centering
\footnotesize

\newcommand\RotText[2]{\rotatebox[origin=c]{90}{\parbox{#1}{\centering#2}}}
\newcommand\NonRotText[1]{\parbox[c]{\linewidth}{\centering#1}}
\renewcommand\tabularxcolumn[1]{m{#1}} %
\newcommand\APat[1]{$\text{AP}_\text{#1}$}

\begin{tabularx}{\linewidth}{lYYYYY}
    \toprule
    Model                                     & CA & mAP   & \APat{50} & \APat{75} & \APat{90} \\
    \midrule
    LLaVA-OV~\cite{li2024llavaonevision} (ZS) &  2.9   & -    & -      & -      & - \\
    MiniCPM-o 2.6 (ZS)                        &  3.0   & -    & -      & -      & - \\
    LMPM~\cite{ding2023mevis}                 &   -    &  6.3 &  13.6  & 5.4    &  1.1    \\
    DsHmp~\cite{he2024dshmp}                  &   -    & 10.4 & 22.3   & 8.9    & 2.0    \\
    VideoLISA~\cite{bai2024onetoken}          &   -    & 7.8  & 17.3    & 6.2   & 1.3 \\
    \midrule
    \ourBaseline{}                            &  3.0   & 16.5 & 32.1   & 15.0   & 3.8    \\
    \bottomrule
    
\end{tabularx}

\vspace{-4pt}
\caption{Benchmark Results on our validation set. Refer to supplementary for test set results. CA: Caption Accuracy}

\label{tab:benchmark_results_test_set}
\end{table}

We provide benchmark results for the test set in Table~\ref{tab:benchmark_results_test_set}. We see that \ourBaseline{} outperforms other baselines and existing task-specific approaches, with the LLama3-8B~\cite{dubey2024llama3} backbone providing the highest performance, which is consistent with the trends seen on validation set.
Compared to the validation set, the test set is more challenging from a segmentation perspective, evident from the lower scores for all methods on the LG-VIS task.

\end{document}